\documentclass[sigconf, nonacm]{acmart}
\usepackage{xspace}
\usepackage{CJKutf8}
\usepackage{float}
\AtBeginDocument{%
  \providecommand\BibTeX{{%
    \normalfont B\kern-0.5em{\scshape i\kern-0.25em b}\kern-0.8em\TeX}}}

\setcopyright{acmcopyright}  
\copyrightyear{2022}
\acmYear{2022}
\acmDOI{TBD}  

\acmConference[Washington, DC '22]{n Proceedings of the 28th ACM SIGKDD Conference on Knowledge Discovery and Data Mining (KDD'22)}{August 14-18, 2022}{Washington, DC}
\acmBooktitle{Washington, DC '22, KDD,
  August 14-18, 2022, Washington, DC}
\acmISBN{TBD}

\begin{document}


\title{A New Generation of Perspective API: \\Efficient Multilingual Character-level Transformers}

\author{Alyssa Lees}
\authornote{Equal contribution, ordered alphabetically.}
\affiliation{%
  \institution{Jigsaw}\city{New York}\country{USA}
}
\email{alyssalees@google.com}

\author{Vinh Q. Tran}
\authornotemark[1]
\affiliation{%
  \institution{Google Research}\city{New York}\country{USA}
}
\email{vqtran@google.com}

\author{Yi Tay}
\authornotemark[1]
\affiliation{%
  \institution{Google Research}\country{Singapore}
}
\email{yitay@google.com}

\author{Jeffrey Sorensen}
\affiliation{%
  \institution{Jigsaw}\city{New York}\country{USA}
}
\email{sorenj@google.com}
\author{Jai Gupta}
\affiliation{%
  \institution{Google Research}\city{Mountain View}\country{USA}
}
\email{jaigupta@google.com}

\author{Donald Metzler}
\affiliation{%
  \institution{Google Research}\city{Mountain View}\country{USA}
}
\email{metzler@google.com}

\author{Lucy Vasserman}
\affiliation{%
  \institution{Jigsaw}\city{New York}\country{USA}
}
\email{lucyvasserman@google.com}


\newcommand{\softmax}{\mathrm{softmax}}
\newcommand{\todo}[1]{\textbf{\color{red}{TODO:}\textit{#1}} \PackageWarning{TODO:}{#1!}}

\begin{abstract}
On the world wide web, toxic content detectors are a crucial line of defense against potentially hateful and offensive messages. As such, building highly effective classifiers that enable a safer internet is an important research area. Moreover, the web is a highly multilingual, cross-cultural community that develops its own lingo over time. As such, it is crucial to develop models that are effective across a diverse range of languages, usages, and styles. In this paper, we present the fundamentals behind the next version of the Perspective API from Google Jigsaw.  At the heart of the approach is a single multilingual token-free Charformer model that is applicable across a range of languages, domains, and tasks. We demonstrate that by forgoing static vocabularies, we gain flexibility across a variety of settings. We additionally outline the techniques employed to make such a byte-level model efficient and feasible for productionization. Through extensive experiments on multilingual toxic comment classification benchmarks derived from real API traffic and evaluation on an array of code-switching, covert toxicity, emoji-based hate, human-readable obfuscation, distribution shift, and bias evaluation settings, we show that our proposed approach outperforms strong baselines. Finally, we present our findings from deploying this system in production.
\end{abstract}





\maketitle

\section{Introduction}
Developing robust and effective content moderation systems is a crucial component for keeping the web safe from abusive users. Offensive, toxic, and harassing content has the potential to significantly harm users along with wide-ranging negative effects on broader society. To this end, building machine learned systems that are able to detect toxic content is a well-established and highly important research area, and an essential component of modern platforms' content moderation workflows, alongside robust human moderator teams. 

Most research in this area has been focused on building specialized models for specific locales, languages, domains, or label distributions \citep{alakrot2018towards,ranasinghe2019brums,mandl2019overview,47349}. 
Thus, it is common practice to build monolingual models for specific languages and/or domains. Given that the web is highly multilingual, multi-cultural, and typographically diverse, monolingual systems are likely to
under-perform in real applications, as they are unable to handle code-switching, 
cross-cultural phenomena, or cross-lingual generalization.

Due to the rigidity of feature-based machine learning models, subword tokenization, and byte-pair encoding \citep{kudo-richardson-2018-sentencepiece} based deep learning models (e.g., BERT \citep{devlin2018bert}), many of these models are not universally applicable across different languages and/or tasks and are considered more or less static once trained. This makes it difficult to apply a single model across a diverse range of languages, domains, and tasks. It also makes it challenging to incrementally train a model on new downstream applications. Furthermore, rigid vocabularies are also vulnerable to common adversaries of the social web - misspellings, emojis and obfuscation,
all of which are techniques that are commonly used for microaggressions and covert attacks \citep{lees-etal-2021-capturing}.

With these challenges in mind, this paper presents a new generation of toxic content classifiers for Jigsaw's Perspective API. We refer to this generation as UTC (\textit{Unified Toxic Content Classification}), centering around a new modeling framework for highly performant and robust toxic content detection. In summary, UTC is a \textbf{single} compact pretrained Charformer-based Transformer \citep{tay2021charformer} that is pretrained on multilingual documents along with comment text from a wide variety of online discussion forums and other sources of user generated content using a sequence to sequence denoising loss \citep{raffel2019exploring}. 

UTC leverages recent novel advances of Learnable Tokenizers as part of the model architecture \citep{tay2021charformer} and is therefore vocabulary- and token-free. While character-level features or modeling approaches have been explored in the context of toxicity detection \citep{kurita2019towards}, our paper proposes the first byte-level pretrained model that remains competitive with (or outperforms) subword models tailored to specific domains or languages. Notably, this vocabulary free property of UTC enables it to be both language agnostic and more robust to domain transfer.

Furthermore, in the design of UTC we address major practical challenges of productionizing character-level Transformers in a latency sensitive, public API setting. We describe the approach in detail in Section \ref{sec:arch} and show the impact of our changes in Table \ref{tab:performance-ablation}. 
The result is a model compact enough to be served in real-life production settings while demonstrating competitive performance versus the winning entries of the 2020 \textit{Jigsaw Multilingual Toxic Comment Classification} Kaggle contest, even though it is >10x more memory (parameter) efficient. Extensive evaluations for model bias \citep{borkan2019} also show that UTC is reasonably unbiased across multiple languages. In summary, the primary contributions of this work can be summarized as follows:
\begin{itemize}
    \item We present Unified Toxic Content Classification (UTC), a modeling framework suitable for efficient character-level multilingual moderation workflows. The proposed UTC framework is comprised of Learnable Tokenizers \citep{tay2021charformer}, Reconfigurable Seq2Seq Transformer architectures, and a new comment-based pre-training scheme. 
    \item We conduct very extensive and rigorous experiments on multiple tasks and benchmark datasets, from both academic settings and sampled from our production traffic. We show that UTC outperforms strong baselines such as a multilingual
    BERT model pretrained on comments and state-of-the-art mT5 models.
   \item For evaluating the robustness and flexibility of the proposed approach, we include benchmarks that specifically test the model's ability to handle code-switching, covert toxicity, emojis, obfuscated text, distribution shifts, and model bias. We show that under all conditions, UTC outperforms or matches strong baselines.
   \item We present the results of our experience deploying UTC into production Perspective API.
\end{itemize}

\section{Related Work}
Although there has been a long history of using machine learning to detect abusive content (e.g., email spam \cite{DADA2019e01802}), research using direct text
classification started in earnest with the introduction of the modest sized
hand labeled
data in \cite{hateoffensive}, \cite{chikashi} and
\cite{waseem-hovy-2016-hateful}. These works
also coincide with the launch of the Workshop on Online Abuse and Harms
\footnote{\url{http://www.workshopononlineabuse.com/}}
which completed its fifth annual meeting.


There has also been a significant amount of criticism regarding the
application of machine learning to conversation moderation,
see \cite{yin} for a recent survey of the issues and challenges. The nature of online
identity and social relationships, and the problems of governance are complex and
involve many interacting entities with overlapping jurisdictions. And despite the
popularity of some shared, labeled test sets, there is little consensus within the
community regarding sampling, annotation standards, annotator recruitment and
training, classifier design, or scoring metrics.

One concern regarding the use of machine learning models for moderation is that
flaws in the training data, whether due to the process used to collect the data, biases held by the annotators, or underlying societal, historical biases, whether intentional or unconscious,
can manifest in models as unintended discriminatory biases. This concern was
raised in \cite{davidson-etal-2019-racial} and \cite{sap-etal-2019-risk}. \cite{jacobs_blodgett} provides a good overview of these concerns. To address
these concerns we employ the techniques of data augmentation suggested in
\cite{dixon2018} and \cite{borkan2019}, and include a cross-language bias analysis to measure the
unintended bias for similar terms across all languages.
It is also worth noting the progress towards a more comprehensive taxonomy
of abusive content has drawn interdisciplinary attention, and systemic
annotation efforts \cite{kennedy_atari} which also inform our work. There has also been criticism regarding current
commercial models' performance on tagging abusive, toxic, or hateful content. This includes
many examples of adversarial perturbations designed to ``fool'' moderation models, such as proposed in \cite{grondahl}. While the present work demonstrates improved
performance against these types of attacks, the task of improving models in these areas is ongoing. 

The English language has dominated research in classifying offensive content,
although there have been numerous publications focusing on other specific languages.
There is still no agreement regarding the efficacy of training multilingual models
versus monolingual models, but for applications with user generated content, not
needing to ask or guess what language is being used has clear advantages.
Two recent examples of similar work are \cite{wang-banko-2021-practical}, which found that
monolingual models do not universally perform better
on sentiment and hate speech classification, and
\cite{song-huang-xiao} which uses model fusion in an
attempt to correct the imbalance in available training
resources.

\newcommand{\modelname}{\textsc{UTC}\xspace}
\newcommand{\bertmodelname}{\textsc{Custom} m\textsc{BERT}\xspace}
\newcommand{\modelnamebase}{\textsc{UTC}\dag\xspace}

\section{UTC: Unified Toxic Content Classification}
\label{sec:arch}
This section introduces UTC, the proposed modeling framework in this paper. 

\begin{figure}[t]
    \centering
    \includegraphics[width=0.3\textwidth]{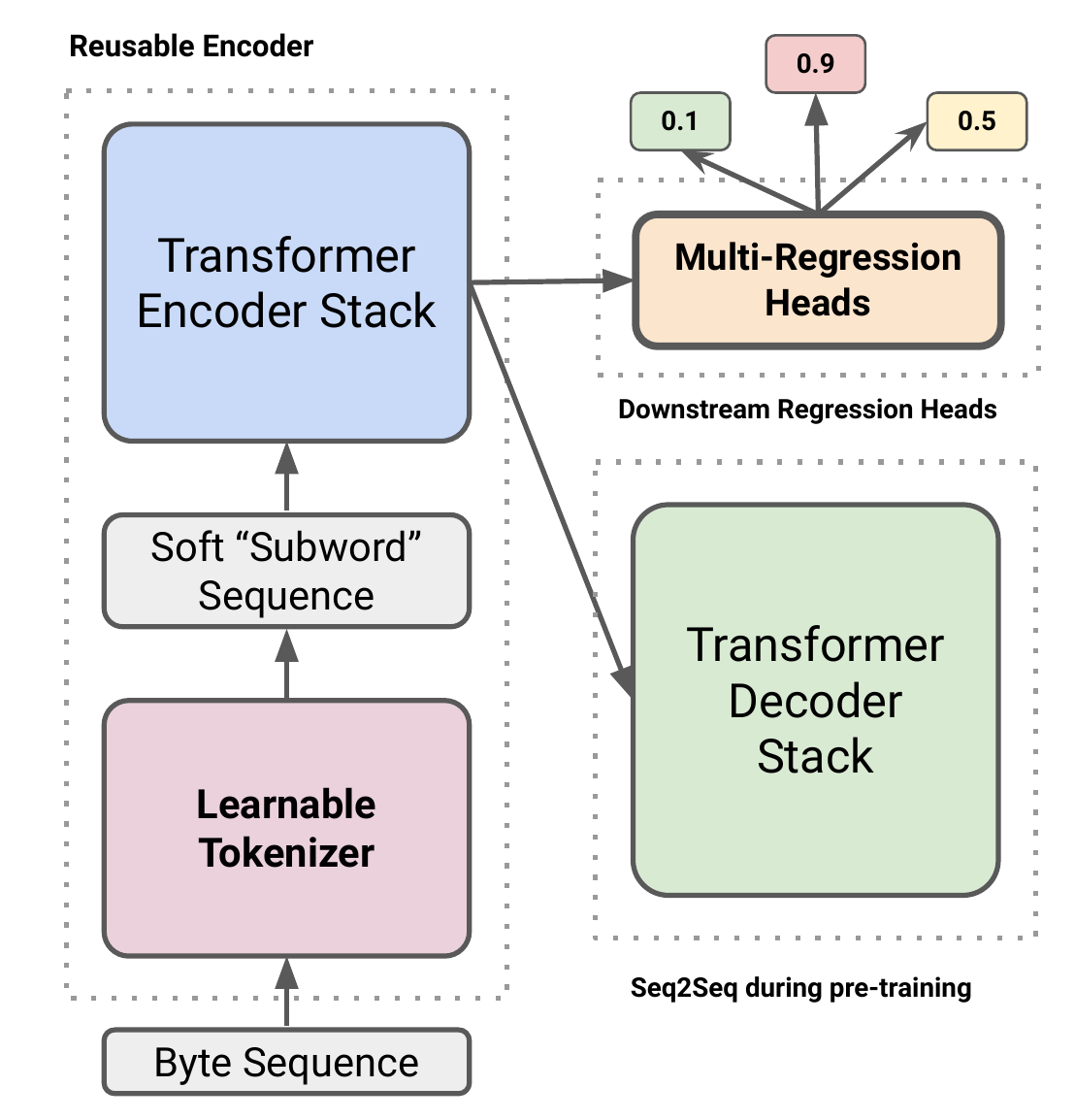}
    \caption{Overview of the UTC architecture.}
    \label{fig:utc-diagram}
\end{figure}


\subsection{Learnable Tokenizer}
The input to our model is a sequence of UTF-8 bytes. After mapping each byte id to an embedding lookup, the input to our model is a tensor $X \in \mathbb{R}^{L_{bytes} \times d_{model}}$ where $L_{bytes}$ is the number of bytes and $d_{model}$ is the number of hidden dimensions. In order to automatically learn subwords in a data-driven fashion, we adopt state-of-the-art Charformer encoders \citep{tay2021charformer}. 

\subsubsection{Learning Latent Subwords Automatically} In this section we review the gradient-based subword tokenization module (GBST) from \citep{tay2021charformer}. GBST dynamically down-samples a sequence of byte embeddings into a sequence of latent subword embeddings in a process that resembles subword tokenization, but can be implemented differentiably. The key idea of GBST is to encourage local composition by performing \textit{position-wise} block scoring. In other words, at every position, we predict a scored list of blocks with each block representing a different size context around the current position. e.g. at a position $i$, we may consider every block of size 1, 2, 3, and 4 intersecting with position $i$. Each block is scored using a block scoring network, parameterized by a simple linear transformation, that maps each candidate subword block embedding into a scalar $\in \mathbb{R}$ that denotes its strength of being included in the final subword composition at the current position. In detail, given a sequence of byte embeddings:

\begin{enumerate}
    \item The model constructs block candidates of varying sizes. Let the maximum possible block size be $M$ and $b$ be the current block size, we use a non-parameterized strided pooling function $F: \mathbb{R}^{b \times d} \rightarrow \mathbb{R}^d$ that projects a subword block consisting of a sequence of byte embeddings $X_{i:i+b} \in \mathbb{R}^{b \times d}$ to a single subword block representation $X_{b, i} \in \mathbb{R}^d$ for block size $b$ at position $i$. When applied across the sequence, we compute a sequence of subword blocks $X_{b}$: \begin{align}X_{b} = [F(X_{i:i+b});F(X_{i+b:i+2b}); \ldots]\end{align} In practice we set $M = 4$ to enumerate blocks sized 1 to 4. Following previous work, since we enumerate blocks here with a stride of $b$ we apply a 1D convolution of size $b+1$ before this enumeration step.
    \item Next we use the block scoring network (a linear transformation) to score every block in each of $X_1, \ldots, X_M$. 
    We then upsample every sequence $X_b$ and their scores back to original sequence length $L_{bytes}$ via repetition. At this point, we have a set of block embeddings $X_{b, i}$ and their scores $p_{b, i}$ for every position $i$ and block size $b$.
    \item We take the softmax of the scores across block size for each position: $P_{i} = \softmax\left([p_{0,i}, p_{1,i}, \cdots, p_{M,i}]\right)$.
    \item We construct the locally composed sequence representation $\hat{X}$ by reducing over the block size dimension. In particular we take the sum of every $X_{b,i}$ at position $i$ weighed by their block score:  $\hat{X}_i = \sum^{M}_b P_{b,i} X_{b,i}$.
    \item Finally $\hat{X}$ is down-sampled by mean pooling.
\end{enumerate}

\noindent We refer interested readers to \cite{tay2021charformer} for fine-grained details.

\subsection{Transformer Stack}
The Transformer stack in our approach accepts latent subwords from the Learnable Tokenizer as an input and the remainder of the Transformer stack remains identical to a standard Transformer model. Transformer architectures are characterized by stacks of self-attention blocks followed by simple feed-forward layers \cite{vaswani}.

\subsection{Reconfigurable Seq2Seq Architecture} Our pretraining utilizes a Seq2Seq (Encoder-Decoder) architecture that is optimized by teacher forcing. In practice, we find this denoising loss to be more effective than encoder-only (BERT-based) pretraining. Intuitively, Seq2Seq based masked language modeling also enables sequential and long-term dependencies to be taken into account in the autoregressive generation process. Our Seq2Seq architecture is reconfigurable, i.e., during certain tasks, we may remove the decoder for specialized regression or classification heads while retaining a universal encoder for all tasks. With this formulation, we can retain a unified encoder across all tasks from shared representation learning. While the T5 model \citep{raffel2019exploring} enables regression problems to be framed in Seq2Seq architectures, we find that adding regression heads is more natural and effective in practice. Moreover, this supports the case where we have multiple labels per input example. Note that the entire UTC Seq2Seq architecture can also be finetuned on downstream classification tasks. 
\subsubsection{Seq2Seq Loss}
During pretraining, our model optimizes the following cross entropy loss: $L = -\sum^{L}_{t=1} \sum^n_{i=1} \log(\pi^t_i) + (1 - y^t_i)\log(1-\pi^t_i)$ where $\pi^t_{i}$ is the prediction of class $i$ at time step $t$ and $y^t_{i}$ is the ground truth label of the class $i$ at time step $t$.
\subsubsection{Multi-Regression Heads and Loss Function}
While the model's main focus is to predict a single value $\in [0,1]$ denoting a \textit{toxicity probability}, our method also generalizes to $k$-way regression to support predicting toxicity subtypes. (For example, if the sample is hateful or obscene). Therefore, for regression tasks, we equip our model with a linear transform that maps the encoder output to a $k_r$ way regression head. This is expressed as: $y_{R} = W_r(\psi(Y'_{out}))$ where $y_{R} \in \mathbb{R}^{k_r}$ and $Y'_{out}$ is the output of the last encoder layer. $\psi$ is a non-parametric or parametric pooling function that maps $\mathbb{R}^{L \times d_{model}} \rightarrow \mathbb{R}^{d_{model}}$ (L is the sequence length)
and $W_r \in \mathbb{R}^{d \times k_r}$ are learnable parameters of the regression head. We adopt a \textit{first pooling} for $\psi$ in similar spirit to BERT's CLS token and include a dummy task prefix token in front of each example following \citep{raffel2019exploring}. Before pooling, we project $Y'_{out}$ using a GeLU MLP layer to the same hidden size, i.e., $d_{model}$ which constitutes the parameters of the pooling layer. For regression head, our model optimizes the sigmoid cross entropy loss.

\subsection{Pretraining}
Our model is pre-trained on an equal mixture of two data sources: Perpsective Pretraining Corpus (PPC) and the mC4 corpus from mT5 \citep{xue2020mt5}. PPC is a proprietary corpus of $\sim$4.6B message and comment texts from a variety of sources including data historically processed by the Perspective API or shared by partners. Note that API clients can enable/disable this data storage via an API flag (\texttt{doNotStore}\footnote{https://developers.perspectiveapi.com/s/about-the-api-methods}). Text in this corpus typically comes from a variety of online forums. We mix the two corpora equally, sampling equally between target languages within the mC4 mixture, while using the natural language distribution in the PPC split.
We pretrain our method using the span-based denoising objective in a Seq2Seq fashion using a \textit{mean} span corruption length of $20$ bytes and a corruption rate of $15\%$. We pretrain for 1M steps and batch size of $128$ sequences, with the maximum length for each sequence set to be $512$ bytes.  

\section{Experimental Settings}
This section provides an overview of our experimental setup.
\subsection{Datasets}
We conduct three categories of experiments: core multilingual toxic comment classification, robustness evaluation, and evaluation of adaptation to new types of toxicity. For core multilingual toxic comment classification we evaluate on both existing public benchmarks (Multilingual Toxic Comments Challenge) as well as a labeled real world dataset derived from live API traffic (Production-Multilingual). For robustness evaluation, we evaluate model performance when faced with code-switching (a subset of the multilingual datasets), obfuscation (obfuscated CivilComments), and distribution shift (zero-shot TweetEval \cite{barbieri-etal-2020-tweeteval} and CivilComments-WILDS \cite{pmlr-v139-koh21a}). We also evaluate the model on an identity term bias task based on \cite{borkan2019}. For adapting to new types of toxicity, we evaluate finetuning performance on Covert Toxicity \cite{lees-etal-2021-capturing}, and Hatemoji \cite{kirk2021}. We refer the reader to Sections \ref{sec:exp-multi}, \ref{sec:robust}, and \ref{sec:adapt} for detailed descriptions of these datasets.

\subsection{Models}
\label{sec:baselines}
This section discusses the details about the major models we use in our experiments.
\begin{itemize}
    \item \textbf{Perspective API} 
    Jigsaw's public API for scoring
    comments for toxicity \cite{Perspective}, prior to this work. It should be noted that many of the languages evaluated in the paper are not currently supported by the Perspective API, and as such Perspective results are omitted in such experiments. 
    \item \textbf{Custom mBERT} We compare with a strong multilingual BERT \citep{devlin2018bert} baseline that has been pretrained on PPC.  The model uses a custom SentencePiece vocabulary of  size 200K, created explicitly from the PPC corpus. We refer to this strong production baseline as \bertmodelname and consider it  representative of a model highly tailored for the domain.
    The baseline model consisted of 768 dimensions, 12 layers, 12 heads, consistent with BERT-base \cite{devlin2018bert}. The pre-training consists of MLM Loss and translation pairs with uniform masking at $15\%$. Pretraining was conducted for 125K steps with batch size of 32K.
    
    \item \textbf{Multilingual T5 (mT5)} - the state-of-the-art for multilingual natural language processing. mT5 is a pretrained T5 \citep{raffel2019exploring} model pre-trained on 100+ languages on the Multilingual C4 corpus.
    \item \textbf{\modelname} and \textbf{\modelnamebase} - our proposed models described in Section \ref{sec:arch}. For the vanilla \modelname model, we use $d_{model}=512$, $d_{ff}=2048, d_{kv}=64, N_{heads}=8$. The number of encoder layers is set to $24$ and the number of decoder layers is set to $6$. For the learned tokenizer, we set the sequence length downsampling rate of $2$, and set the convolution filter size to $5$. The standard \modelname is approximately 102M parameters when deployed in downstream applications. This model size was considered to ensure a fast serving latency. We also consider a larger (but still servable) \modelnamebase model that is approximately $268M$ parameters where  $d_{model}=768$, $d_{ff}=3072, d_{kv}=64, N_{heads}=12$ and number of encoder layers is set to $28$. We denote this model as \modelnamebase.  
\end{itemize}
Fine-grained details on each specific baseline can be found in each individual experiment section. Please see Appendix \ref{sec:details} for additional reproduction details.

\begin{table*}[h!]
    \centering
    \small
    \begin{tabular}{l|c|cccccccccccc|c}
    \hline
      Model   & Params & Ar & Cs & En & Hi-Latn & Id & Ja & Ko & Nl & Pl & Pt & Ru & Zh & Avg \\
      \hline
      Perspective API & - & - &- & .974 &- &- &- &- &- &- &- & \textbf{.907} & - & -  \\
      \bertmodelname & 235M & .762 & .881 & \textbf{.982} & .832 & .812 & .649 & .855 & .842 & .853 & .878 &  .803 & .925 & .840 \\  
      \hline
    mT5$_{small}$ & 148M & .896 & .913 & .969 & .962 & .761 & .881 & .846 & .726 & .866 & .856 & .880 & .976  & .878\\ 
    mT5$_{base}$  & 278M &  .900 & .925  & .973 & .967  & .791 & .887 & .874 & .934 & .881  & .850 & .888 & \textbf{.997}  & .906 \\ 
    \hline
       \modelname & 102M & .899 & .925 & .977 & .954 & .794 & .867 & \textbf{.938} & .940 & .892 & .864 & .896  & .975 & .910  \\
      \modelnamebase &  268M & \textbf{.908} & \textbf{.934} & .977 & \textbf{.968} & \textbf{.819} & \textbf{.896} & .916 & \textbf{.947} & \textbf{.896} & \textbf{.888} & \textbf{.907} & .974 & \textbf{.919} \\ 
       \hline
    \end{tabular}
    \caption{Experimental results on the Production-Multilingual dataset. We report AUC-ROC scores.}
    \label{tab:jigsaw-multi}
    \vspace{-2em}
\end{table*}

\section{Experiments: Multilingual}
\label{sec:exp-multi}
In this section we report the core multilingual toxic comment classification results of the work. With the exception of Perspective API, we finetune and evaluate all models outlined in Section \ref{sec:baselines}, on each dataset. using a batch size of 512 until convergence.

\subsection{Production-Multilingual}
\label{sec:jigsaw-multi}

\subsubsection{Dataset}

A proprietary internal multilingual toxic comment classification training and evaluation set. This dataset is derived from live traffic that is sent to the production Perspective API (with \texttt{doNotStore} flag set to false), translated to multiple languages and is exclusively labeled offline by human annotators for toxicity. Given the nature of this dataset, this task represents the performance of our models in production, and is the most important metric by which we compare models. The training sets cover the languages of AR, CS, DE, EN, ES, FR, HI, HI-Latn, ID, IT, JA, KO, NL, PL, PT, RU, SV, ZH along with some low prevalence examples in a few other languages. The training data is ~38 million records and is not balanced across languages with a heavy skew towards EN. The evaluation sets are limited to the languages covered in the experimental results : AR, CS, EN, HI-Latn, ID, JA, KO, NL, PL, PT, RU, ZH. This narrower set focuses in on languages where Perspective did not already have a production quality model (at the time of experimentation), plus English where significant Perspective usage comes from. The evaluation sets are roughly balanced in volume across languages and comprise ~1.3 million records.

\subsubsection{Results} Table \ref{tab:jigsaw-multi} reports results on the Production-Multilingual dataset. Overall, \modelname outperformed all baselines, with a small \modelname model outperforming mT5$_{base}$, a model with more than twice the size w.r.t. number of parameters. While \modelname does not perform as strongly on English as  \bertmodelname, the main advantage of \modelname is observed in many non-English languages. 

\subsection{Multilingual Toxic Comments Challenge}
\subsubsection{Dataset} In this section, we report experimental results on the public dataset
featured in the Jigsaw
Multilingual Toxic Comments Challenge (JMTCC) hosted by Kaggle. The competition was held in 2020 and comprises of 6 languages besides English: Spanish, French, Italian, Portuguese, Russian, and Turkish. While the evaluation data was multilingual, only English data was provided in the training set. Hence, it was common for participants to make use of translation data to augment the training set. We train our models on the translated data that was shared in the Kaggle discussion forums.

\subsubsection{Compared Baselines} Aside from the baselines in Section \ref{sec:baselines}, we also report results from the winners of the Jigsaw Multilingual Toxic Comment Classification Kaggle competition, although it is worth noting that our goal is to develop \textbf{single} standalone models that can feasibly be deployed in production. Meanwhile, the top Jigsaw Multilingual Toxic Comment Classification Kaggle submissions often involved aggressive ensembling, score scaling techniques etc, that are highly infeasible in practice. Nevertheless, we believe it is beneficial to evaluate how well our standalone single model fares compared to a strong highly engineered upper bound. Based on our interpretation of the Kaggle champion's entry, we estimate the number of model parameters to be $>5B$ given that they ensemble multiple XLM large models (at least 300M parameters each) along with monolingual models. 
\begin{table}[t]
    \centering
    \small
    \begin{tabular}{l|cc}
    \hline
      Model   &  \# Params &AUC-ROC  \\
      \hline
    Kaggle \# 1 & $\approx$>5B$^\ast$ & \textbf{.9536} \\
    \hline
        Perspective API & - & .8770* \\
       \bertmodelname & 235M & .9104 \\
    \hline
    mT5$_{small}$ & 148M  & .9156 \\ 
    mT5$_{base}$ & 278M & .9239   \\ 
    \hline
    \modelname & 102M & .9194 \\ 
    \modelnamebase & 268M & \textbf{.9367} \\ 
    \hline
    \end{tabular}
    \caption{Results on Jigsaw Multilingual Toxic Comments Challenge (JMTCC). *Turkish is not supported by Perspective API, and is omitted from this result.}
    \label{tab:kaggle}
    \vspace{-2em}
\end{table}

\subsubsection{Results} Table \ref{tab:kaggle} reports results on the JMTCC dataset. Our results show that our best \modelnamebase achieves $0.9367$ AUC-ROC, outperforming all considered \textit{single model} baselines, especially a strong state-of-the-art mT5 baseline. Notably, this result is only slightly worse than the top performing Kaggle \#1 result which comprises XLM-Roberta ensembles, pseudo labelling and other commonly used techniques. We consider the result achieved by \modelnamebase to be pretty compelling, given that this is a single model that can actually be used in production applications. 

\section{Experiments: Robustness}
\label{sec:robust}
In this section we do no additional training and evaluate the fine-tuned models from Section \ref{sec:jigsaw-multi} to evaluate the robustness of our proposed methods.

\subsection{Code-Switching}
The Code-Switching eval sets aim to identify theoretically more difficult multilingual user comments. Both bespoke evaluation datasets below are constructed by filtering the parent superset with the same criteria for multilingual comment identification: test examples are restricted to those where $2$ or more languages are present. Samples are included if and only if $>=25\%$ of the example content is identified to be in each of 2 or more languages using a language detection model.\footnote{https://github.com/google/cld3}. We use two subsets for code-switching based on Production-Multilingual and JMTCC datasets. Details on the breakdown of these code-switching datasets can be found in the supplemental material.





\subsubsection{Compared Baselines}
The same baseline models evaluated on the Production-Multilingual dataset were employed, including \bertmodelname, the comment domain multilingual BERT model, $mT5_{base}$ and $mT5_{small}$.  Similarly, we also included an evaluation using the public Perspective API \cite{Perspective}. It should be noted that not all of the languages included in the code-switching datasets are listed as supported by Perspective and as such Perspective may be disadvantaged in this experiment. To specify a language for Perspective API, we use a language detection model to identify the primary language for each code-switch example. If the language is not supported by Perspective API, then we default to English. 

\begin{table}[t]
    \centering
    \small
    \begin{tabular}{l|ccc}
    \hline
      Model   &  \#Params & JMTCC-CS & Production-CS   \\
      \hline
    Perspective API & - & .7516 & .7163 \\
    \bertmodelname & 235M & .9243 & .8106\\
    \hline
     mT5$_{small}$ & 148M  & .9289 & .8661\\ 
    mT5$_{base}$ & 278M &  .9393 & .8730  \\ 
    \hline
    \modelname & 102M & .9191 & .8755 \\ 
     \modelnamebase & 268M & \textbf{.9446} & \textbf{.9023} \\ 
    \hline
    \end{tabular}
     \caption{Experiments on Code-Switching. }
    \label{tab:kaggle_cs}
    \vspace{-2em}
\end{table}

\subsubsection{Results} Table \ref{tab:kaggle_cs} reports our experimental results on code-switching evaluation sets. On both JMTCC-CS and Production-CS, \modelnamebase outperforms the best, outperforming the mT5$_{base}$ baseline. In general we find that an off-the-shelf mT5 model also substantially outperforms the \bertmodelname model. Finally, we note that the Perspective API performs poorly since prior to this work, models were separately trained on individual languages and as such they are not well equipped to handle multilingual code-switching. One limitation of this experiment is the dominance of English and Latin based languages in the code-switching evaluation sets. We suspect that the byte level vocabulary in our Charformer model, is advantageous for understanding with character based languages especially in code-switching tasks. The preliminary results give evidence to this conclusion.


    

\subsection{Human-Readable Obfuscation}
\label{sec:obfuscation}
One common technique used to bypass toxicity classification models is to intentionally misspell words in a fashion that is understood by human readers, yet obfuscated to machine learning models \cite{grondahl}. Even though our proposed models are not explicitly trained on these types of adversarial examples, in this section we run zero-shot experiments on synthetically obfuscated data to evaluate model robustness in this area.

\subsubsection{Dataset} We construct a synthetically obfuscated variant of the Civil Comments \cite{borkan2019} test set. As the dataset is in English only, we manually construct a dictionary of valid substitutions for every letter in the English alphabet. Then for every alphabetical character in each example, we replace the character by a substitute with some probability, which we call the character obfuscation rate. If a character is chosen, then a substitute for the character is chosen uniformly at random from the list of valid substitutes for the character. Substitutes may be any other character or string that may still be readable as original character, e.g. "a" may be substituted with "4", "@", or "/\textbackslash". Valid substitutions for vowels also include "*" or an empty string, which effectively removes information from the sequence. See Figure \ref{fig:obfu-dictionary} for the comprehensive dictionary of valid substitutions, and Figure \ref{fig:obfu-examples} for examples of text at various character obfuscation rates. Finally, please note that the construction of this dataset is not meant to comprehensively capture a realistic distribution of adversarial examples against toxicity classification models. Instead, we aim to create a controlled and sufficiently challenging dataset to serve as a point of evaluation between models.

\subsubsection{Compared Baselines} We evaluate the zero-shot performance of Perspective API (prior to this work), \bertmodelname, mT5-small, and \modelname (102M param.) on obfuscated Civil Comments, sweeping the character obfuscation rate from 0 to 50\% in increments of 10\%. \bertmodelname, mT5, and \modelname are the same fine-tuned models from Table \ref{tab:jigsaw-multi}.  No additional training was done for these experiments. 

\begin{figure}[t]
\includegraphics[width=0.38\textwidth]{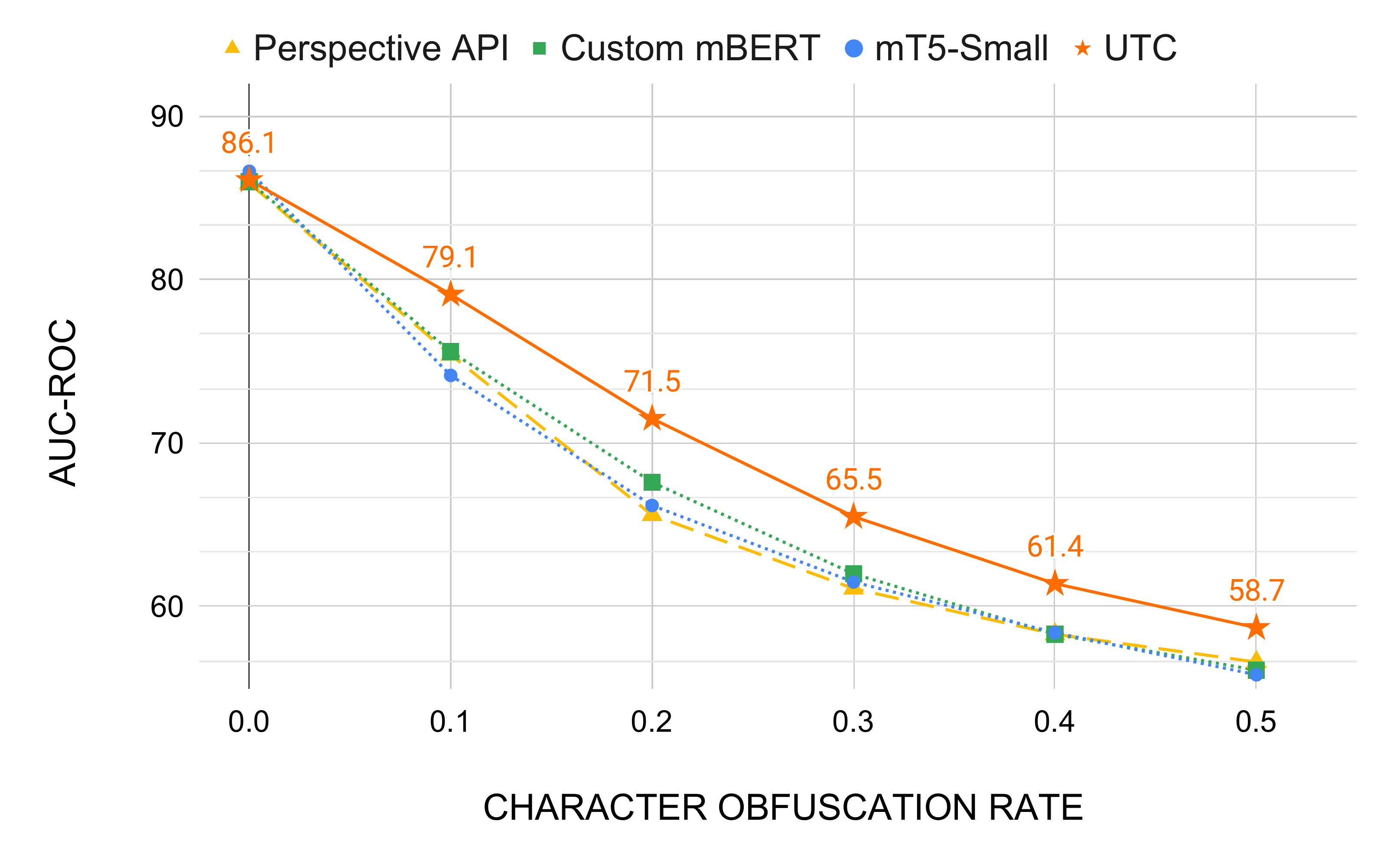}
\vspace{-1em}
\caption{Zero-shot AUC-ROC on English-only CivilComments with 0-50\% obfuscation rate.}
\label{fig:obfuscation}
\vspace{-1em}
\end{figure}

\subsubsection{Results} Figure \ref{fig:obfuscation} plots the performance of all models across character obfuscation rates. In this experiment we observe that while all models have similar zero-shot performance when there is no obfuscation applied to the English only dataset, \modelname outperforms every baseline at every other obfuscation rate greater than 0. Albeit, all models do decay in performance as obfuscation rate increases -- however, this is expected as the models rarely see this type of obfuscation during training, and are not fine-tuned on any additional obfuscated data. This result echos a similar finding by previous byte-level models \cite{byt52021}. One natural question to ask might be: if fine-tuned, are the models able to learn to adapt to this type of obfuscation? As an additional result, we found that when fine-tuned on a 30\% obfuscated version of Civil Comments, UTC was able to fully recover performance to 86.0 AUC-ROC, while mT5-small recovers to only 84.5 AUC-ROC (-2.1pt from the unobfuscated zero-shot result.) This result illustrates the value of the UTC inductive bias on this particular task.

\subsection{Distribution Shifts}
Toxic content appears on many different surfaces in different forms, targeting many different types of people. In this section we evaluate the performance of our model on two different setups. In the first setup with TweetEval, we evaluate performance on a task with a different labeling process and domain focus. In the second setup, we evaluate the performance of the model on subpopulation shift using  CivilComments-WILDS.

\subsubsection{TweetEval Hate Classification}
\paragraph{Dataset} We evaluate our models on the TweetEval hate content classification test split \cite{barbieri-etal-2020-tweeteval}, which is taken from the SemEval2019 Hateval challenge \cite{basile-etal-2019-semeval}. The task is to predict whether a given tweet contains hateful language targeted against any of two communities: women and immigrants. This task differs from our UTC pre-training and fine-tuning as it is purely Tweet focused and has been labeled to a different standard (i.e. hateful language.) As this is zero-shot evaluation, we do not do any additional fine-tuning for this experiment.

\paragraph{Compared Baselines} We evaluate our same baselines and models as Section \ref{sec:obfuscation}: Perspective API, \bertmodelname, mT5, and \modelname. For mT5 and \modelname we evaluate both small and base sized versions. Additionally, we compare our zero-shot results against the current state-of-the-art for the task: RoBERTa Retrained and BERTweet. Both are English-only RoBERTa-based models which have been extensively pre-trained on a large corpus of English tweets, as well as finetuned on the corresponding TweetEval hate classification training set. 

\begin{table}[t]
    \small
    \centering
    \begin{tabular}{l|cc}
    \hline
      Model   &  \# Params & \textsc{Macro. F1}  \\
      \hline
    RoBERTa-Retrained (Finetuned avg. of 3) \cite{barbieri-etal-2020-tweeteval} & 125M  & 52.3 \\ 
    RoBERTa-Retrained (Finetuned best) \cite{barbieri-etal-2020-tweeteval} & 125M & 55.5 \\ 
    BERTweet (Finetuned best) \cite{nguyen-etal-2020-bertweet} & 125M & 56.4 \\
    \hline
    Perspective API (Zero-shot) & - & 52.0 \\
    \bertmodelname (Zero-shot) & 235M & 51.8 \\
    mT5$_{small}$ (Zero-shot) & 148M & 53.9\\
    mT5$_{base}$ (Zero-shot) & 278M & 57.7 \\
    \hline
    \modelname (Zero-shot) & 102M & 53.3 \\ 
    \modelnamebase (Zero-shot) & 268M & 55.1 \\
    \hline
    \end{tabular}
    \caption{Performance on TweetEval hate classification.}
    \label{tab:tweethate}
        \vspace{-2em}
\end{table}

\subsubsection{Results} Table \ref{tab:tweethate} reports performance on TweetEval hate classification. All zero-shot baselines which were finetuned on Production-Multilingual data showed strong performance in this experiment, with multilingual mT5 and \modelname in particular performing on par with an English-only RoBERTa model that had seen additional pretraining on a large Twitter corpus and finetuned on TweetEval hate classification training data. Higher results previously reported by \cite{barbieri-etal-2020-tweeteval} and \cite{nguyen-etal-2020-bertweet} are only observed in "best run" performance where the model saw favorable variance. This experiment demonstrates the effectiveness of our methods in training domain shift robust toxicity classification models.

\subsubsection{CivilComments-WILDS}
\paragraph{Dataset} Introduced in \cite{pmlr-v139-koh21a}, this dataset augments the Civil Comments dataset with various demographic identities referenced in each example. The goal of this dataset is to evaluate for subpopulation shift: a setting where the model sees all domains (i.e. demographic identities) during evaluation as it does during training, but in different proportions. In particular, for CivilComments-WILDS a model is trained on all demographic identities available in CivilComments, but is evaluated on a single identity at a time -- an extreme subpopulation shift. This is repeated individually for each subpopulation, with the aim of maximizing performance on the worst performing subpopulation. Following the original work, we perform our analysis on 8 demographic identities male, female, LGBTQ, Christian, Muslim, other religions, Black, and White. We report accuracy on the complete test split and the worst accuracy from the 8 subpopulations and the gap between the two to show that our model performs better on these subpopulation shifts. \cite{pmlr-v139-koh21a}, showed the existence of a significant gap between the average in-distribution accuracy and the worst subpopulation accuracy. We additionally report this gap for our own evaluated models.

\paragraph{Compared Baselines}
We evaluate small-sized mT5 and \modelname models in this setting from Table \ref{sec:jigsaw-multi}. We compare to the highest performing DistilBERT results from \cite{pmlr-v139-koh21a} with respect to both average overall accuracy and worst-group accuracy (DistilBERT using empirical risk minimization, ERM, and group distributionally robust optimization, Group DRO, respectively.) As our models were multilingually fine-tuned, while DistilBERT fine-tuned on only English Civil Comments, for a fair comparison in this subpopulation shift setting our mT5 and \modelname models are additionally fine-tuned on Civil Comments before evaluation. We do not use any robust optimization techniques when fine-tuning our models.

\begin{table}[t]
    \small
    \centering
    \resizebox{\columnwidth}{!}{%
    \begin{tabular}{l|cccc}
    \hline
      Model   &  \#Params & \textsc{Avg Acc}  & \textsc{Worst Acc} & \textsc{Gap}\\
      \hline
    DistilBERT ERM \cite{pmlr-v139-koh21a} & 66M  & 92.2 & 56.0 & 36.2 \\ 
    DistilBERT DRO \cite{pmlr-v139-koh21a} & 66M & 89.9 & 70.0 & 19.9 \\
    \hline
    mT5$_{small}$ & 148M & 94.0 & 81.8 & 12.2\\
    \hline
    \modelname & 102M & 94.2 & 82.6 & 11.6 \\ 
    \hline
    \end{tabular}
    }
    \caption{Accuracy on CivilComments-WILDS dataset.}
    \label{tab:wilds}
    \vspace{-3em}
\end{table}

\paragraph{Results} Both mT5-small and \modelname significantly outperform the baseline results from prior work on all metrics. Our models perform better overall and have almost half to a third smaller of a gap between average and worst group performance than DistilBERT with robust optimization techniques. Although the exact source of this gain is unclear, we posit that the extended amount of multilingual pre-training, pre-finetuning and greater model size may play a significant role here.

\subsection{Identity Term Bias}
Borkan et al. \cite{borkan2019} outlined nuanced bias metrics, to be used in addition to overall model metrics such as AUC-ROC and \cite{dixon2018}, introduced synthetic, templated datasets for identifying unintended bias in toxicity models. Here we use these tools to evaluate our models for identity term bias.

\subsubsection{Dataset.} We use a new multilingual version of the synthetic template bias evaluation data set, publicly released in 2021\footnote{https://medium.com/jigsaw/identifying-machine-learning-bias-with-updated-data-sets-7c36d6063a2c and https://github.com/conversationai/unintended-ml-bias-analysis/tree/2021-refresh}. The examples in this dataset are generated from predefined templates with slots where different words (e.g. identity terms, adjectives, verbs, etc.) can be substituted for related terms in order to test for performance with regards to various subgroups (identities). To obtain a multilingual dataset for each of the 12 target languages we rely on a team of expert native speakers to construct these templates. The final generated dataset consists of $\sim2M$ examples and has a balanced class distribution. Following \cite{borkan2019} we report Subgroup AUC, Background Positive Subgroup Negative (BPSN) AUC, and Background Negative Subgroup Positive (BNSP) AUC for all subgroup-language combinations. Please see \cite{borkan2019} for precise definitions of these metrics.

\subsubsection{Results.} We evaluate model bias for \modelname with mT5-small serving as a baseline comparison. Both models remain the same as from Table \ref{tab:jigsaw-multi}. We visualize the results in Figure \ref{fig:mt5-bias} and \ref{fig:bias} for all language splits. Note that to generate these visualizations we aggregate results with their corresponding English identity term (results with no corresponding English term are not shown here). Overall, the metrics remain strong at $>$.7 across languages, with \modelname performing stronger on more subgroup-language combinations than mT5. As both models are finetuned on the same dataset, we see that the UTC inductive bias may play a role here. However, some subgroup-language combinations still require additional work. For example, there are some terms in Korean and Japanese that demonstrate unwanted bias, such as the BPSN AUC metric for the term \textit{homosexual} where
\begin{CJK}{UTF8}{}
\CJKfamily{mj}
동성애자
\end{CJK}
, and 
\begin{CJK}{UTF8}{min}
同性愛者
\end{CJK} have values $\leq$.5. This suggests that the non-toxic templates containing the identity term are yielding false positives, or rather the term is correlated with toxicity.  As such, further explicit debiasing efforts are still needed.

\section{Experiments: Adapting}
\label{sec:adapt}
In this section we demonstrate that the fine-tuned checkpoints from Section \ref{sec:jigsaw-multi} are highly adaptable to new types of toxicity by further fine-tuning our models on new challenging tasks.

\subsection{Covert Toxicity}
We conduct experiments on Covert Toxicity \citep{lees-etal-2021-capturing}, a task of distinguishing if a piece of text contains nuanced
toxicity such as microaggressions. We compare with Toxic-BERT and Covert-BERT baselines reported in  \citep{lees-etal-2021-capturing}. We also finetune a monolingual (English only)
T5 and mT5 base model as strong baselines.


\subsubsection{Results} Table \ref{tab:covert} reports results on the CovertToxicity task. We show that \modelname and \modelnamebase achieves very competitive results outperforming both Toxic-BERT and Covert-BERT baselines. On this task, the performance of \modelnamebase is competitive to mT5$_{base}$. Interestingly, the multilingual models (mT5 and \modelname model) outperform the specialized monolingual English T5 model. 

\subsection{Emoji-based Hate}

\subsubsection{Dataset} The English-only Hatemoji dataset comprises of two splits: \textsc{HatemojiCheck} and \textsc{HatemojiTrain}. \textsc{HatemojiCheck} is a manually constructed labeled test suite of 3,930 short-form statements and whether they use emoji-based hateful language. 

\subsubsection{Compared Baselines} \cite{kirk2021} showed that fine-tuning on \textsc{HatemojiTrain} greatly improves performance on \textsc{HatemojiCheck}. Following this, we evaluate our two highest performing models from Section \ref{sec:jigsaw-multi}, mT5 and \modelname, by further fine-tuning them on \textsc{HatemojiTrain} then evaluating on \textsc{HatemojiCheck}. Note that these checkpoints have already been finetuned on Production-Multilingual. We use the validation split of \textsc{HatemojiTrain} to pick the best checkpoint for this additional fine-tuning. We additionally compare these results to the best results reported in \cite{kirk2021}, which is an English-only  DeBERTa model optimized for the task. 


\begin{table}[t]
    \small
    \centering
    \begin{tabular}{l|c}
    \hline
    \small
    Model     &  AUC-ROC\\
    \hline
    Toxic-BERT & .520 \\
    Covert-BERT & .590 \\ 
    Monolingual T5$_{base}$ & .599 \\ 
    mT5$_{base}$ & \textbf{.607} \\
       \hline
       \modelname & .604 \\ 
         \modelnamebase &  \textbf{.607} \\ 
         \hline
    \end{tabular}
    \caption{Results on CovertToxicity.}
    \label{tab:covert}
    \vspace{-2em}
\end{table}

\begin{table}[t]
    \small
    \centering
    \begin{tabular}{l|ccc}
    \hline
      Model   &  \# Params & \textsc{Acc.} & \textsc{F1}  \\
      \hline
    Kirk et al. \cite{kirk2021} & 140M & 87.9 & 91.0 \\
    mT5$_{small}$ & 148M  & 86.0 & 89.6 \\
    mT5$_{base}$ & 278M  & 86.8 & 90.5 \\ 

    \hline
    \modelname &  102M & \textbf{90.0} & \textbf{92.8}\\ 
    \modelnamebase & 268M & \textbf{90.8} & \textbf{93.3} \\
    \hline
    \end{tabular}
    \caption{Performance on English-only \textsc{HatemojiCheck}.}
    \label{tab:hatemoji}
    \vspace{-2em}
\end{table}

\subsubsection{Results} Table \ref{tab:hatemoji} reports results on \textsc{HatemojiCheck}. Even though \modelname is multilingual, when finetuned \modelname outperforms the best performing model from \cite{kirk2021} by a significant margin. We posit that this gain may be attributed to the learned tokenizer, which may effectively update during finetuning to adapt to parsing emojis. On the other hand mT5$_{Small}$, under-performs the Kirk et al. baseline.

\section{Deployment Results}
On December 9th, 2021 Jigsaw launched support for 10 new languages\footnote{Languages launched with UTC: Arabic, Chinese (Simplified), Czech, Dutch, Indonesian, Japanese, Korean, Polish, Hindi, and Hinglish (a mix of English and Hindi transliterated using Latin characters)} in the Perspective API \cite{jigsawblog}, powered by \modelnamebase (Table \ref{tab:jigsaw-multi}), making the model available publicly \footnote{https://developers.perspectiveapi.com/s/about-the-api-attributes-and-languages} (Note that languages\footnote{Languages not yet using UTC: English, French, German, Italian, Portuguese, Russian, and Spanish} previously available within Perspective API were not impacted by this launch). UTC dramatically increased Perspective's capabilities, as it is the first model to reach our production standards for these 10 languages. All previous production candidates (CNNs and BERT-based architectures) had low overall performance, low performance on bias evaluations, or were too slow to serve for real-time usage. 

The model was deployed smoothly with no operational issues, and as of writing this paper, the model averages $\sim$15 QPS and $\sim$200ms median latency (for the 10 newly launched languages only). In our load testing, we have observed that the smaller \modelname (102M) model can achieve 45ms median latency at 1K QPS on a single TPUv2 chip with batching. We anticipate our production latency improving further as load increases as our serving infrastructure does not have to wait to accumulate requests for batching. In the future, we hope to migrate to using the smaller and faster \modelname (102M) model, as well as explore further performance improvements. We also plan to transition the rest of the languages Perspective serves to UTC over time, as well as expand to additional new languages. 

\begin{table}[t]
    \small
    \centering
    \begin{tabular}{l|cc}
    \hline
      Model   &  \# Params & Finetune Steps/s \\
      \hline
    Byte-level T5 Base & 200M & 18.3 \\
    \;\;\;\;+ Charformer & 134M  & 26.5 \\
    \;\;\;\;\;\;\;\;+ Regression Head (\modelname) & 102M  & 32.9 \\ 
    \;\;\;\;\;\;\;\;\;\;\;\;+ Increased Scale (\modelnamebase) & 268M  & 15.0 \\ 

    \end{tabular}
    \caption{Architecture ablation for Production-Multilingual finetuning speed on 64 TPUv3 chips (batch size 128, 512 byte input length.) These values correlate with serving latency.}
    \label{tab:performance-ablation}
    \vspace{-3em}
\end{table}

Overall, we consider our results impressive for a \textit{byte-level} Transformer model of this size. We attribute the majority of this performance to the sequence length downsampling done by Charformer, as well as our removal of the decoder during finetuning, effectively reducing both our sequence length and depth dimensions by half respectively. We include an ablation study for the speed of our model with and without these modifications in Table \ref{tab:performance-ablation}. Performance may also be attributed to forgoing tokenization, a process that is hard to parallelize, in favor of Charformer GBST, which can run on specialized hardware (TPU). In addition to quality and performance, there are engineering advantages to our approach. We have found that having one model to support multiple languages significantly simplifies the maintenance of our service as now there are fewer models to maintain. Additionally, our experience forgoing tokenization echoes that of previous literature -- we find that preparing models for production is simplified as we no longer need to coordinate model checkpoints with matching vocabularies. Given these results, we are looking forward to expand the usage of this approach in the future.

\section{Conclusion}
This paper presents Jigsaw's new generation of toxic comment classification models, which is currently deployed in production for 10 new languages in the Perspective API. We outline our approach in applying state-of-the-art token-free Charformer to the problem of toxic comment classification and the efficiency techniques we take to enable serving such a byte-level model in production. Through rigorous experiments on real-world and academic benchmarks we demonstrate the effectiveness our approach.

\bibliographystyle{ACM-Reference-Format}
\bibliography{main}

\newpage
\appendix
\section{Reproduction Details}
\label{sec:details}
\subsubsection{Implementation} Our \modelname model is implemented in Mesh TensorFlow\footnote{\url{https://github.com/tensorflow/mesh}} \citep{shazeer2018mesh}, a wrapper over TensorFlow API that enables distributed model parallelism, along with the T5 library\footnote{\url{https://github.com/google-research/text-to-text-transfer-transformer}}. For Charformer \citep{tay2021charformer}, we use the official implementation\footnote{\url{https://github.com/google-research/google-research/tree/master/charformer}} released by the authors. The overarching model architecture follows the T5.1.1 setup using T5-styled relative attention biases instead of position embeddings. 


\subsubsection{Optimization and Training Details} This section describes the general setup for our pretraining and finetuning experiments. Dataset specific details are deferred to respective sections. For both pretraining and finetuning, we use the Adafactor optimizer \citep{shazeer2018adafactor}. During pretraining, we use a learning rate equal to the inverse square root of the current training step following \citep{raffel2019exploring}. Finetuning is performed using a fixed constant learning rate of $10^{-3}$. We apply a dropout of $0.1$ during finetuning. Pretraining is conducted with $64$ TPU-v3 chips and finetuning is typically conducted with $16$ TPU-v3 chips. Pretraining generally takes about 3-4 days to complete. 

\subsubsection{Reproducibility} Our model is currently available via the production Perspective API\footnote{https://developers.perspectiveapi.com/s/docs} for Arabic, Chinese (Simplified), Czech, Dutch, Hindi, Hinglish (a mix of Hindi and English), Indonesian, Japanese, Korean, Polish, and Russian for the "TOXICITY" attribute. Access to the model for English is also released under the "TOXICITY\_EXPERIMENTAL" attribute. Even though the interface for the API requires specification of a language, all requests are routed to a single \modelnamebase model.

\subsubsection{Dataset Details} Here we include figures to further provide some details about selected datasets used.

\begin{table}[H]
    \small
    \centering
    \begin{tabular}{l|c|c}
    \hline
        Language & Prevalence & Proportion \\
        \hline
        en & 1658 & $99\%$ \\
        pt & 474 & $28\%$ \\
        es & 356 & $21\%$ \\
        it & 353 & $21\%$ \\
        fr & 307 & $18\%$ \\
        ru & 147 & $9\%$ \\
        ar, bg, co, de, el, hi, ka, ja, tr, zh & 33 & $<2\%$ \\
        
        \hline
        total & 1664 & 100 \\
    \hline
    
    \end{tabular}
    \caption{JMTCC Code-switching Eval: Language breakdowns of filtered code-switching examples. Note: as an example contains multiple languages, the total does not correspond to the sum of the columns here.}
    \label{tab:kaggle_cs_breakdown}
\end{table}
\begin{table}[H]
    \small
    \centering
    \begin{tabular}{l|c|c}
    \hline
        Language & Prevalence & Proportion \\
        \hline
        en & 31101 & $97\%$ \\
        pt & 12499 & $39\%$ \\
        hi-Latn & 8520 & $26\%$ \\
        id & 6589 & $20\%$ \\
        ru & 636 & $2\%$ \\
        ar & 601 & $2\%$ \\
        es & 417 & $1\%$ \\
        nl & 396 & $1\%$ \\
        de & 353 & $1\%$ \\
        pl & 300 & $1\%$ \\
        fr, ja, it, zh, hi, da, ur, cs, cv, fy, ko, + & 2971 & $11\%$\\
        
        \hline
        total & 32196 & 100 \\
    \hline
    
    \end{tabular}
    \caption{Production-Multilingual Code-switching Eval: Language breakdowns of code-switching examples. Note: as an example contains multiple languages, the total does not correspond to the sum of the columns here.}
    \label{tab:perspective_cs_breakdown}
\end{table}
\begin{figure}[H]
\includegraphics[width=.5\textwidth]{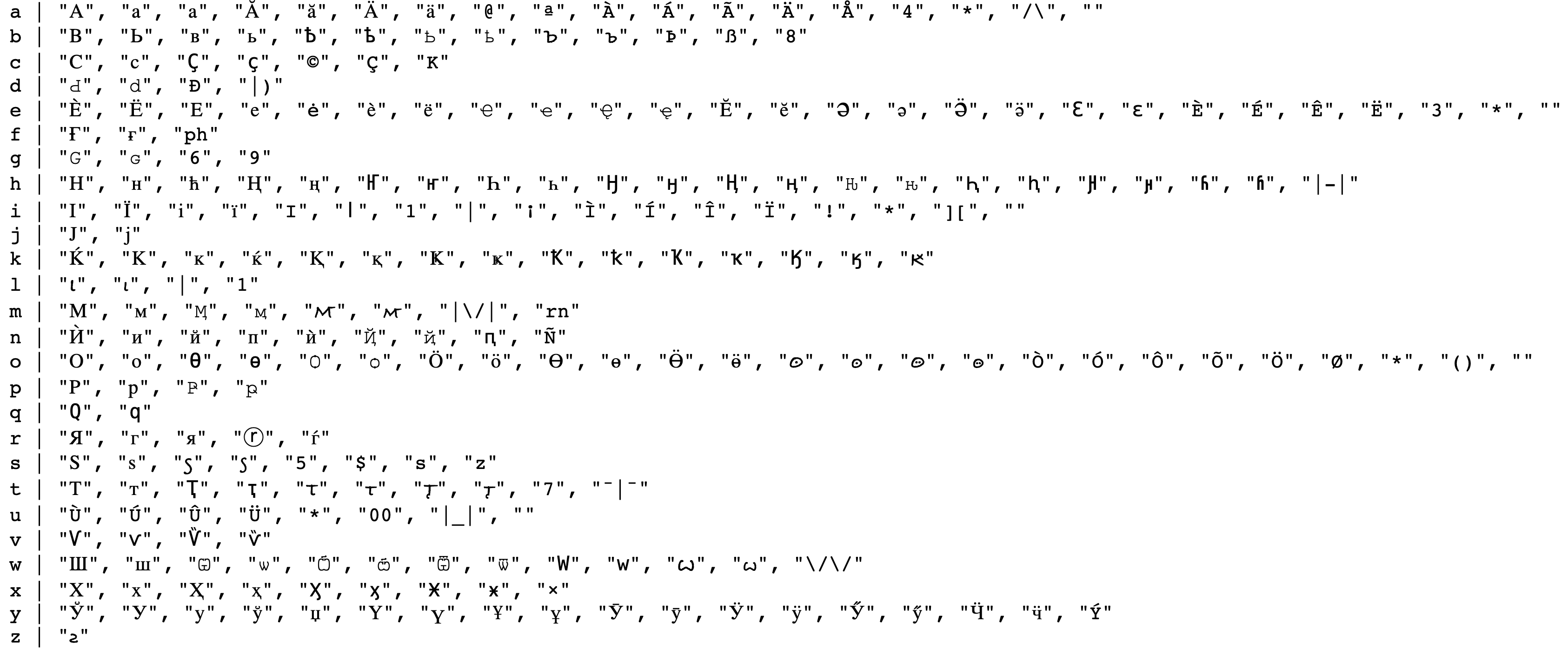}
\caption{Full list of substitutions used for obfuscation experiments in Section \ref{sec:obfuscation}.}
\label{fig:obfu-dictionary}
\end{figure}
\begin{figure}[H]
\includegraphics[width=.5\textwidth]{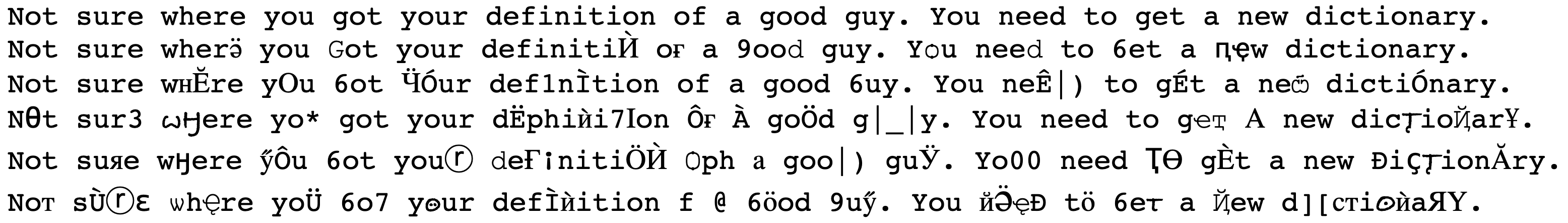}
\caption{Examples of obfuscation of a sentence sampled from Civil Comments, for character obfuscation rate from 0 (top) to 50\% (bottom).}
\label{fig:obfu-examples}
\end{figure}
\begin{figure*}[htbp!]
\includegraphics[width=0.8\textwidth]{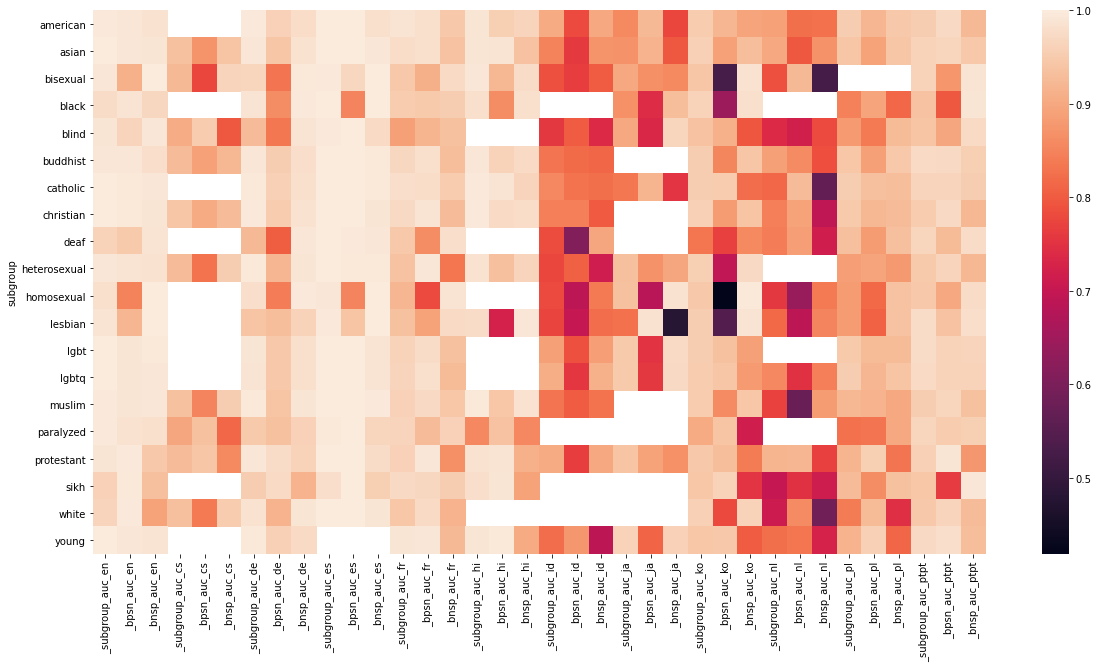}
\caption{mT5 Unintended Bias Metrics, AUC, BPSN, BNSP per Language on template identity eval set for a subset of localized identity terms}
\label{fig:mt5-bias}
\end{figure*}
\begin{figure*}[htbp!]
\includegraphics[width=0.8\textwidth]{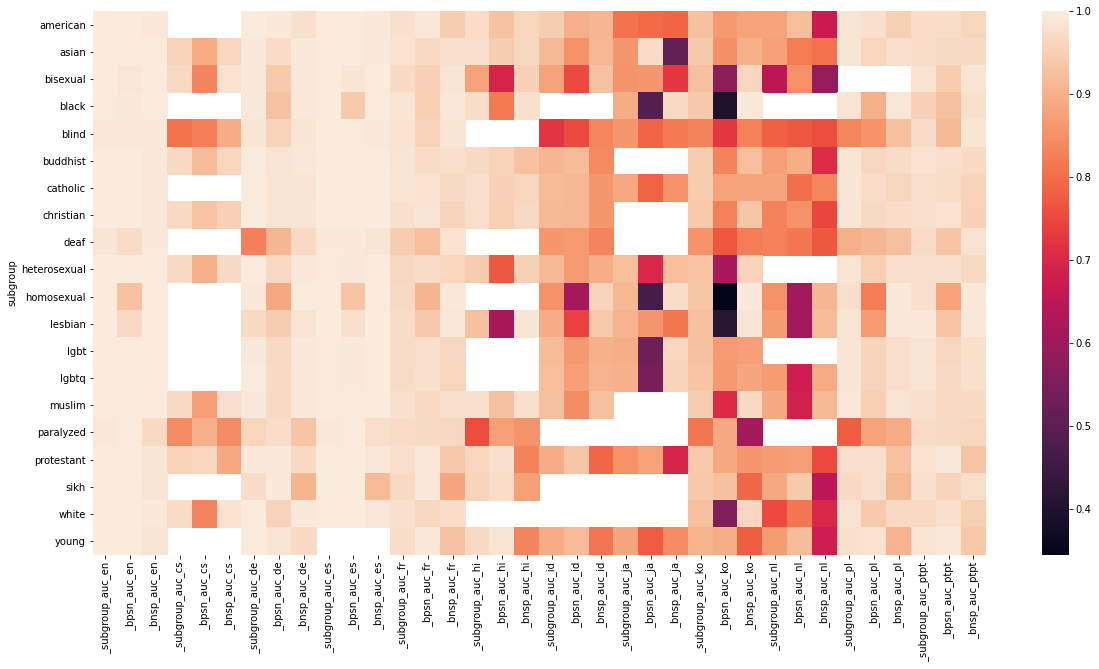}
\caption{\modelname Unintended Bias Metrics, AUC, BPSN, BNSP per Language on template identity eval set for a subset of localized identity terms}
\label{fig:bias}
\end{figure*}
\end{document}